\documentclass[conference]{IEEEtran}

\usepackage{graphicx}
\usepackage{amsmath,amssymb}
\usepackage{booktabs}
\usepackage{tabularx}
\usepackage{multirow}
\usepackage{microtype}
\usepackage{xcolor}
\usepackage{url}
\usepackage[hidelinks]{hyperref}
\usepackage[caption=false,font=footnotesize]{subfig}

\hypersetup{
  colorlinks=true,
  linkcolor=blue,
  citecolor=blue,
  urlcolor=blue
}

\title{Evidence-Grounded Multimodal Knowledge Graph Construction for Multi-Lecture Educational Reasoning}

\author{
\IEEEauthorblockN{Sahil Al Farib\IEEEauthorrefmark{1},
Momota Ahsana Meem\IEEEauthorrefmark{2},
Sheikh Redwanul Islam\IEEEauthorrefmark{3}, and
Md. Tanvir Raihan\IEEEauthorrefmark{4}}
\IEEEauthorblockA{Department of Computer Science \& Engineering, United International University, Dhaka, Bangladesh\\
\IEEEauthorrefmark{1}\texttt{sfarib222186@bscse.uiu.ac.bd},
\IEEEauthorrefmark{2}\texttt{mmeem222254@bscse.uiu.ac.bd}\\
\IEEEauthorrefmark{3}\texttt{sislam222142@bscse.uiu.ac.bd},
\IEEEauthorrefmark{4}\texttt{tanvir@cse.uiu.ac.bd}\\
ORCID: \IEEEauthorrefmark{1}0009-0006-5911-8144;
\IEEEauthorrefmark{2}0009-0003-0334-2567}
}

\begin{document}
\maketitle

\begin{abstract}
Lecture videos distribute knowledge across speech, slide text, diagrams, equations, and presentation order, which transcript-only retrieval does not fully preserve. This paper presents an evidence-grounded multimodal pipeline that transcribes lectures, selects semantic anchors, applies optical character recognition (OCR), and uses a vision-language model to extract only concepts and typed relationships supported by transcript, OCR, or visual evidence. Mentions are validated and canonicalized into a provenance-rich knowledge graph. On three neural-network lectures, the pipeline processed 3,118 frames, 756 transcript segments, and 559 anchors. It retained 1,022 concept and 312 relationship mentions, yielding 172 canonical concepts and 282 relationships with 90.38\% endpoint coverage. A preliminary three-question retrieval test achieved 100\% top-1 and top-3 accuracy and 100\% mean top-5 recall. The contribution is an auditable construction method rather than a state-of-the-art performance claim.
\end{abstract}


\section{Introduction}
Educational concepts may be spoken, written on slides, drawn in diagrams, represented in equations, and revisited across lectures. Students integrate these signals to connect definitions, examples, and prerequisites. Automated lecture question answering therefore needs a representation that retains concepts, relationships, evidence, and temporal context rather than only flat transcript chunks.

Retrieval-augmented generation (RAG) grounds language models in external context, but conventional transcript RAG is weakest on explicit dependencies, concept evolution, and visually communicated information. A knowledge graph (KG) can instead represent concepts as nodes and typed relations as edges while retaining the lecture, timestamp, frame, and evidence quotation that justify each extraction. Such auditability is important in education because unsupported claims can mislead learners and instructors need to inspect or correct extracted knowledge.

We present an end-to-end pipeline combining timestamped automatic speech recognition (ASR), high-recall anchor selection, OCR, grounded vision-language extraction, validation, canonicalization, KG construction, and GraphRAG-style retrieval. The design constraint is simple: a concept or relationship is promoted only when tied to lecture evidence. The study uses three conceptually connected 3Blue1Brown lectures on neural networks, gradient descent, and backpropagation.

The contributions are: (1) a multimodal lecture-to-KG workflow; (2) an extraction schema retaining evidence, confidence, modality, lecture, timestamp, frame, and anchor identifiers; (3) a provenance-rich graph with 172 canonical concepts and 282 relationships; (4) hybrid semantic, lexical, fuzzy, and evidence-aware retrieval; and (5) reproducible figures and tables derived from the final run.

\section{Related Work}

\subsection{Retrieval-Augmented and Graph-Structured Generation}
Lewis \emph{et al.} introduced RAG for knowledge-intensive natural-language processing and showed the value of conditioning generation on retrieved documents \cite{lewis2020rag}. Subsequent surveys have organized the fast-growing RAG literature around indexing, retrieval, and generation stages and have consistently identified hallucination and outdated internal knowledge as central failure modes that external retrieval is meant to address \cite{wu2024rag}. GraphRAG extends this motivation with entity, relation, and community indexes for corpus-level questions that flat chunk retrieval can miss \cite{edge2024graphrag}. More recent work generalizes GraphRAG-style retrieval to non-textual evidence: multimodal RAG surveys report that grounding generation in images, video, and other modalities alongside text further reduces hallucination relative to text-only retrieval, particularly when visual and textual cues are jointly necessary to answer a question \cite{mei2025multimodalrag}. Our work applies graph-structured retrieval to multimodal lectures and, unlike most RAG systems that retrieve opaque passages, retains source evidence for every extracted item so that each answer can be traced to a specific transcript span, OCR string, or frame.

\subsection{Multimodal Knowledge Graph Construction}
Multimodal knowledge graphs (MMKGs) integrate text, images, and other modalities into a single structured representation to support cross-modal reasoning. Zhu \emph{et al.} survey MMKG construction and completion built primarily from text--image pairs and highlight symbol grounding---linking a textual entity to its supporting visual evidence---as a central open problem \cite{zhu2022mmkg}. Large language models have since been used to automate parts of this pipeline. A recent survey of LLM-empowered KG construction describes how vision-language models can be cascaded to translate visual features into text before extraction and catalogs LLM-driven entity-fusion and canonicalization methods that merge duplicate mentions using embedding similarity rather than hand-written rules \cite{bian2025llmkg}. This strategy is conceptually similar to the alias, fuzzy-matching, and embedding-based merging used in our canonicalization stage.

Video-native MMKGs such as Kuaipedia link entities to short-video evidence at scale but are built for open-domain, single-clip content rather than multi-lecture instructional material with typed pedagogical relations such as \texttt{prerequisite\_of} or \texttt{computed\_by} \cite{pan2022kuaipedia}. Our pipeline differs by constraining extraction to a fixed, education-oriented relation vocabulary and requiring every mention to cite the transcript, OCR, or visual evidence that supports it before entering the graph.

\subsection{Educational Knowledge Graphs}
Educational knowledge graphs (EduKGs) structure course concepts and their relationships to support curriculum design, prerequisite discovery, and personalized learning. A systematic review of KG construction and application in education finds that most existing EduKGs are built from static, already-structured materials such as textbooks, syllabi, or course descriptions, and identifies automatic construction directly from raw instructional content as an open challenge \cite{abus2024edukg}. Work on the CourseMapper platform compares top-down (ontology-first) and bottom-up (extraction-first) strategies for automatically building EduKGs from lecture slides and later reports an optimized pipeline that improves concept-extraction accuracy and processing efficiency over the initial approach \cite{ain2025topdown,ain2025optimized}. These efforts operate on slide text or transcripts alone. In contrast, our pipeline treats video frames, OCR output, and transcript segments as three simultaneous evidence sources for the same anchor, allowing it to capture diagram- and equation-only content that slide- or transcript-only EduKG pipelines cannot see.

\subsection{Speech, Vision, and Text Recognition Components}
Lecture processing depends on robust ASR. Whisper demonstrated strong generalization from large-scale weak supervision \cite{radford2022whisper}; we use Faster-Whisper \texttt{large-v3} and align each timestamped segment with a primary frame. For joint interpretation of transcript, slide text, diagrams, and equations, we use Qwen2.5-VL \cite{bai2025qwen} under a constrained JSON extraction prompt rather than as an unconstrained generator. Because slide and diagram text is useful only when read correctly, OCR acts as a second visual-evidence channel. Surveys of deep-learning-based scene-text recognition describe how cluttered backgrounds, varied fonts, and non-frontal capture make this setting harder than scanned-document OCR \cite{chen2021text}, matching the noisy, camera-captured slide frames encountered in lecture video. Canonical graph text is embedded with BGE-large English \cite{bge2023model}.

\subsection{Video Question Answering}
Because the graph is intended to answer lecture questions, our retrieval stage relates to video question answering (VideoQA), which pairs video featurization, question featurization, and joint embedding to produce an answer directly from raw video \cite{jeshmol2024videoqa}. VideoQA systems typically reason over a fixed clip without an explicit intermediate knowledge representation, so they cannot easily aggregate evidence for a single concept across multiple, separately recorded lectures or expose why a particular answer was produced. Our graph-grounded retrieval first resolves a question to canonical concepts and their one-hop neighborhood, then generates an answer only from retrieved definitions, relationships, and evidence. This keeps reasoning inspectable and reusable across lectures rather than tied to a single video pass. Unlike educational KGs manually curated from textbooks or structured materials, the proposed graph is built from raw videos while preserving inspectable multimodal provenance.

\section{Method}

\subsection{Pipeline and Alignment}
\begin{figure}[t]
  \centering
  \includegraphics[width=\columnwidth]{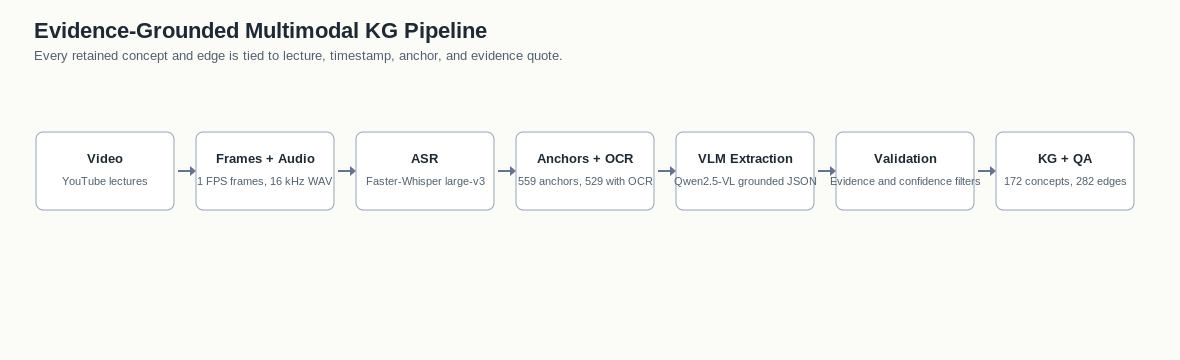}
  \caption{Evidence-grounded multimodal KG construction pipeline.}
  \label{fig:pipeline}
\end{figure}

Fig.~\ref{fig:pipeline} summarizes the workflow. Videos are downloaded, frames are sampled at 1 frame/s, and audio is converted to a 16-kHz mono waveform. Faster-Whisper produces segments with start time, end time, and text; the segment midpoint identifies its primary frame. This synchronization lets transcript and visual evidence share a temporal neighborhood.

\subsection{Semantic Anchors and OCR}
Running a vision-language model (VLM) on every frame is expensive and redundant. A recall-oriented selector combines visual-change, transcript-keyword, relationship-cue, and first-mention scores. It targets approximately 18\% of frames with temporal spacing to suppress near-duplicates. Each anchor receives transcript context from 15~s before through 22~s after its timestamp. EasyOCR processes the anchor frame and contributes slide labels, diagram annotations, symbols, and equations that may not be spoken.

\subsection{Grounded Extraction and Validation}
For every anchor, Qwen2.5-VL receives the frame, transcript window, OCR text, and locally derived candidate terms. It returns strict JSON arrays for concepts and relationships; empty arrays are permitted. A concept records name, definition, evidence quotation, source modality, and confidence. A relationship records source concept, target concept, one of eight permitted types (\texttt{prerequisite\_of}, \texttt{component\_of}, \texttt{uses}, \texttt{optimizes}, \texttt{computed\_by}, \texttt{example\_of}, \texttt{contrasts\_with}, or \texttt{related\_to}), evidence quotation, source modality, and confidence.

Validation removes missing or low-information fields, unsupported evidence, invalid relation types, and confidence below 0.55. Transcript- and OCR-sourced claims must occur in the evidence pool; visual-only concepts require stronger confidence. Validation improves auditability but does not by itself guarantee factual correctness.

\subsection{Canonicalization and Graph Construction}
Validated mentions are merged using aliases, normalized and fuzzy string matching, token overlap, and embedding similarity. Each canonical node retains its identifier, display name, aliases, definitions, mention list, lecture coverage, evidence count, and average confidence. Relationship endpoints are mapped only after concept deduplication; this order prevents otherwise valid edges from being lost when raw endpoint names differ from canonical names.

The result is a NetworkX \texttt{MultiDiGraph}. Nodes are canonical concepts and edges are typed relationship instances. Edge metadata includes relation type, lecture, timestamp, anchor, evidence quotation, and confidence. Multiple edge instances between the same nodes are allowed when supported by different evidence.

\subsection{Graph-Grounded Retrieval and QA}
Canonical concept text combines name, definition, aliases, and evidence snippets and is embedded using BGE-large English. The retrieval score combines semantic similarity, exact-name and alias matches, fuzzy similarity, and an evidence-count prior. The top six concepts are expanded to a one-hop subgraph. The answer prompt receives only formatted definitions, relationships, and evidence and requests lecture identifiers and timestamps when available.

\section{Experimental Setup}

\subsection{Dataset and Configuration}
\begin{figure}[t]
  \centering
  \includegraphics[width=\columnwidth]{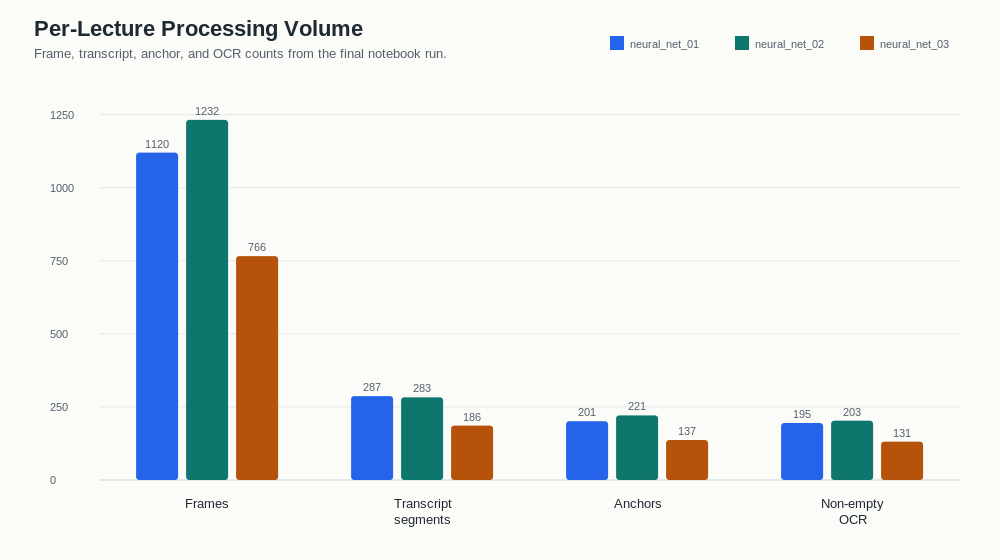}
  \caption{Per-lecture frame, transcript-segment, anchor, and OCR counts.}
  \label{fig:lecturecounts}
\end{figure}

The evaluation covers three lectures on neural networks, gradient descent, and backpropagation. Table~\ref{tab:dataset} reports processing and OCR coverage; Fig.~\ref{fig:lecturecounts} visualizes per-lecture volume. The run used Faster-Whisper \texttt{large-v3}, EasyOCR, Qwen2.5-VL-7B-Instruct, BGE-large-en-v1.5 embeddings, 1-frame/s sampling, an 18\% anchor target, and 0.55 minimum concept and relation confidence.

\begin{table}[t]
\caption{Dataset, Anchor, and OCR Summary}
\label{tab:dataset}
\centering
\small
\begin{tabular}{lrrrrr}
\toprule
Lecture & Frames & Segs. & Words & Anchors & OCR \\
\midrule
Neural networks & 1120 & 287 & 3277 & 201 & 195 \\
Gradient descent & 1232 & 283 & 3698 & 221 & 203 \\
Backpropagation & 766 & 186 & 2222 & 137 & 131 \\
\midrule
Total & 3118 & 756 & 9197 & 559 & 529 \\
\bottomrule
\end{tabular}
\end{table}

\begin{table}[t]
\caption{Models and Processing Configuration}
\label{tab:configuration}
\centering
\small
\begin{tabularx}{\columnwidth}{@{}lX@{}}
\toprule
Component & Model or setting \\
\midrule
ASR; OCR & Faster-Whisper \texttt{large-v3}; EasyOCR \\
VLM; embeddings & Qwen2.5-VL-7B-Instruct; BGE-large-en-v1.5 \\
Graph; sampling & NetworkX \texttt{MultiDiGraph}; 1 frame/s \\
Anchor target; window & 18\%; 15 s before and 22 s after \\
Confidence thresholds & 0.55 for concepts and relationships \\
\bottomrule
\end{tabularx}
\end{table}

The overall anchor and non-empty OCR rates were 17.93\% and 94.63\%, respectively. Reported metrics cover frames, transcripts, anchors, OCR, raw and validated mentions, canonical nodes, graph edges, endpoint coverage, evidence density, and retrieval. Retrieval uses only three seed questions and is a sanity check, not a final benchmark.

\section{Results}

\subsection{Extraction Yield}
\begin{figure}[t]
  \centering
  \includegraphics[width=\columnwidth]{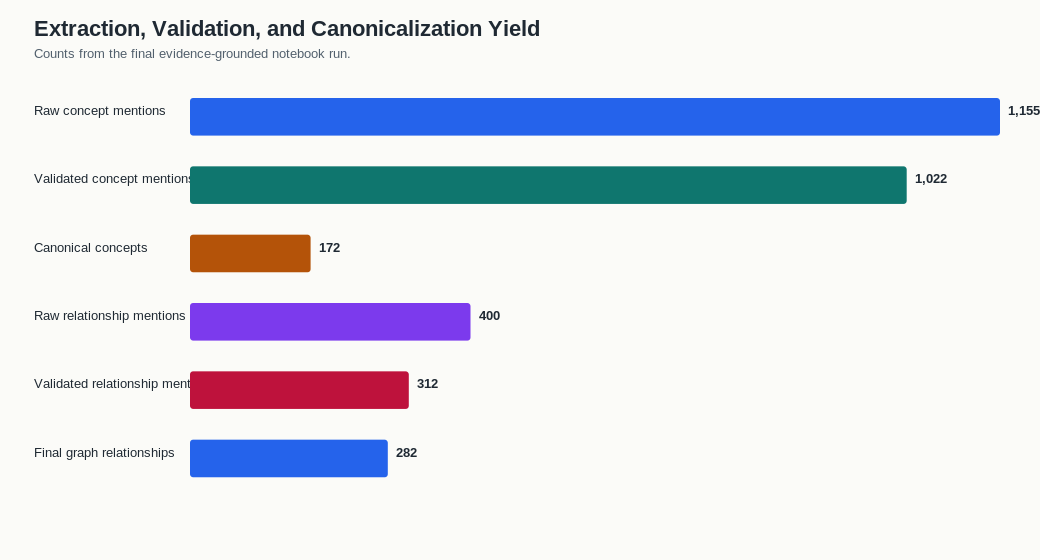}
  \caption{Yield from raw mentions to validated and canonical graph items.}
  \label{fig:funnel}
\end{figure}

The VLM returned 1,155 raw concept and 400 raw relationship mentions. Evidence and confidence checks retained 1,022 concepts (88.48\%) and 312 relationships (78.00\%). Canonicalization produced 172 concepts, and endpoint mapping retained 282 edges (90.38\% of validated relationships), as summarized in Table~\ref{tab:yield} and Fig.~\ref{fig:funnel}. The reduction from 1,022 mentions to 172 nodes is expected because course concepts recur across anchors and lectures.

\begin{table}[t]
\caption{Extraction and Graph-Construction Yield}
\label{tab:yield}
\centering
\small
\begin{tabular}{lrr}
\toprule
Stage & Count & Retention/coverage \\
\midrule
Raw concept mentions & 1155 & -- \\
Validated concept mentions & 1022 & 88.48\% \\
Canonical concepts & 172 & 16.83\% \\
Raw relationship mentions & 400 & -- \\
Validated relationship mentions & 312 & 78.00\% \\
Final graph relationships & 282 & 90.38\% \\
\bottomrule
\end{tabular}
\end{table}

\subsection{Knowledge Graph Statistics}
\begin{figure*}[t]
  \centering
  \includegraphics[width=0.92\textwidth]{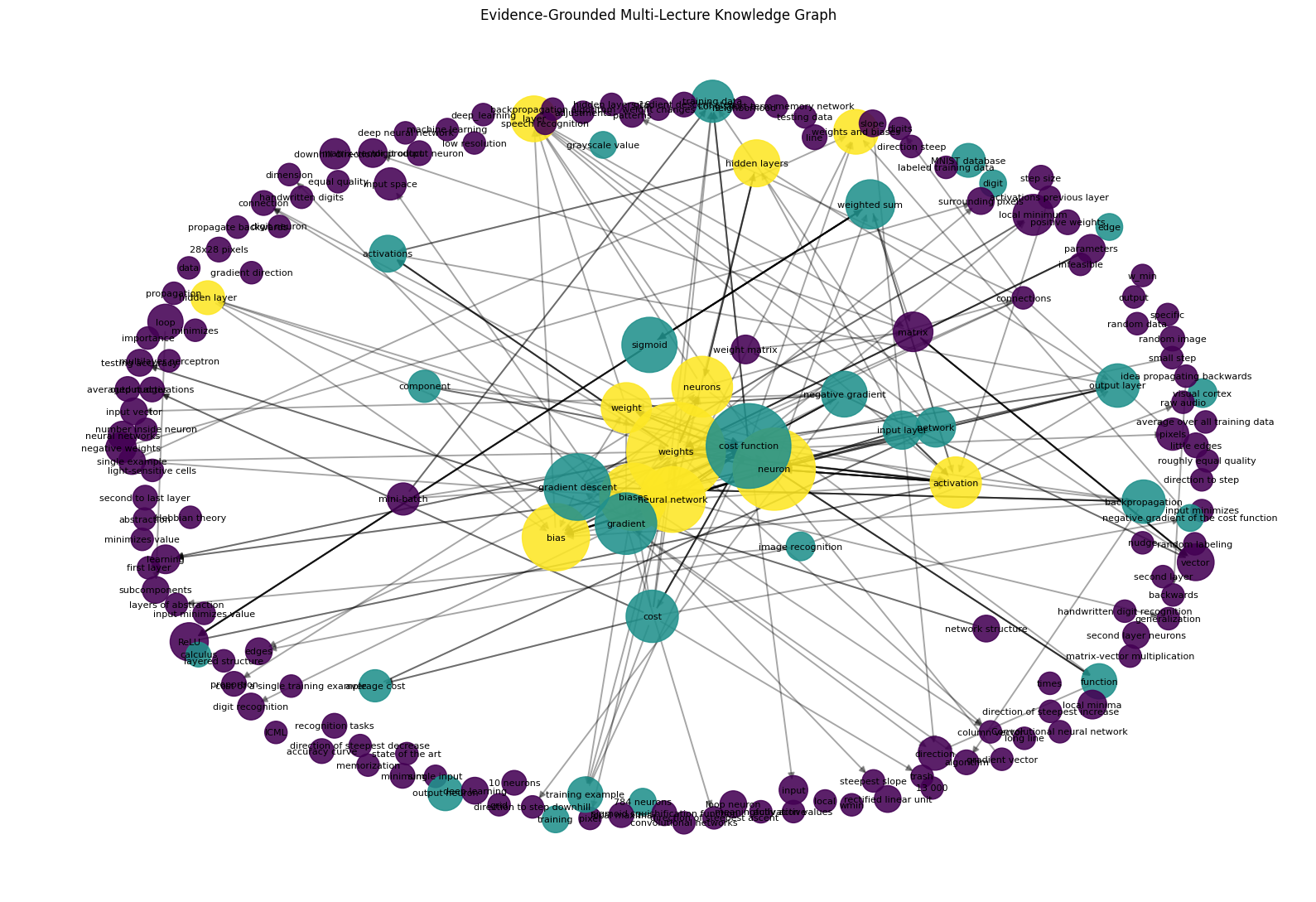}
  \caption{Constructed educational knowledge graph.}
  \label{fig:graph}
\end{figure*}

The graph contains 172 canonical concepts and 282 relationships, with 5.94 evidence mentions per concept on average. Fig.~\ref{fig:graph} shows the constructed graph. Canonical concepts are represented as nodes and extracted typed relationships as edges. Dense clusters indicate neural-network concepts recurring across lectures and anchors, whereas peripheral nodes represent lower-frequency or localized evidence.

\begin{table}[t]
\caption{Graph and Starter-Retrieval Summary}
\label{tab:summary}
\centering
\small
\begin{tabular}{lr}
\toprule
Metric & Value \\
\midrule
Canonical concepts & 172 \\
Final graph relationships & 282 \\
Relationship endpoint coverage & 90.38\% \\
Average evidence per concept & 5.94 \\
Starter retrieval top-1 accuracy & 100.00\% \\
Starter retrieval top-3 accuracy & 100.00\% \\
Starter retrieval mean top-5 recall & 100.00\% \\
\bottomrule
\end{tabular}
\end{table}

\subsection{Concept Evidence and Retrieval}
\begin{figure}[t]
  \centering
  \includegraphics[width=\columnwidth]{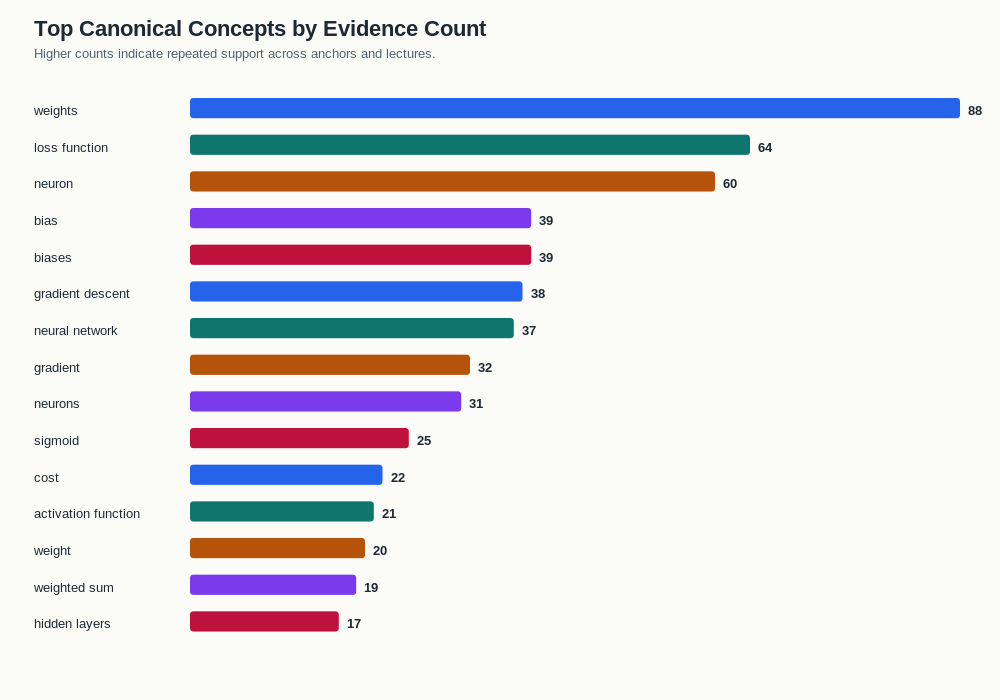}
  \caption{Concepts with the most evidence-backed mentions.}
  \label{fig:concepts}
\end{figure}

Central training concepts dominate the evidence distribution. Table~\ref{tab:concepts} lists the 15 most frequent nodes, and Fig.~\ref{fig:concepts} provides the corresponding visualization. Remaining singular/plural variants reveal conservative but incomplete entity resolution.

\begin{table}[t]
\caption{Canonical Concepts with the Most Evidence}
\label{tab:concepts}
\centering
\small
\resizebox{\columnwidth}{!}{%
\begin{tabular}{lrrlrr}
\toprule
Concept & Ev. & Conf. & Concept & Ev. & Conf. \\
\midrule
weights & 88 & .852 & neurons & 31 & .839 \\
loss function & 64 & .848 & sigmoid & 25 & .836 \\
neuron & 60 & .852 & cost & 22 & .845 \\
bias & 39 & .862 & activation function & 21 & .848 \\
biases & 39 & .854 & weight & 20 & .870 \\
gradient descent & 38 & .839 & weighted sum & 19 & .858 \\
neural network & 37 & .854 & hidden layers & 17 & .829 \\
gradient & 32 & .825 & & & \\
\bottomrule
\end{tabular}
}
\end{table}

\begin{figure}[t]
  \centering
  \includegraphics[width=\columnwidth]{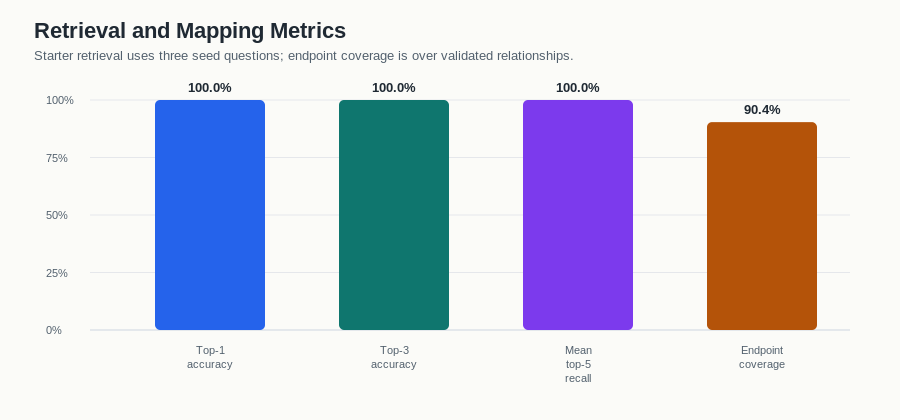}
  \caption{Starter retrieval metrics and relationship endpoint coverage.}
  \label{fig:metrics}
\end{figure}

All three seed questions placed their target concept first and within the top three, with mean top-5 recall of 1.00 (Table~\ref{tab:retrieval}). Fig.~\ref{fig:metrics} contrasts these preliminary scores with endpoint coverage. Definition, relation, prerequisite, and cross-lecture trials retrieved relevant concepts and produced grounded descriptions, but generated answers sometimes added correct background knowledge not explicitly supported by retrieved evidence.

\begin{figure*}[t]
  \centering
  \includegraphics[width=0.96\textwidth]{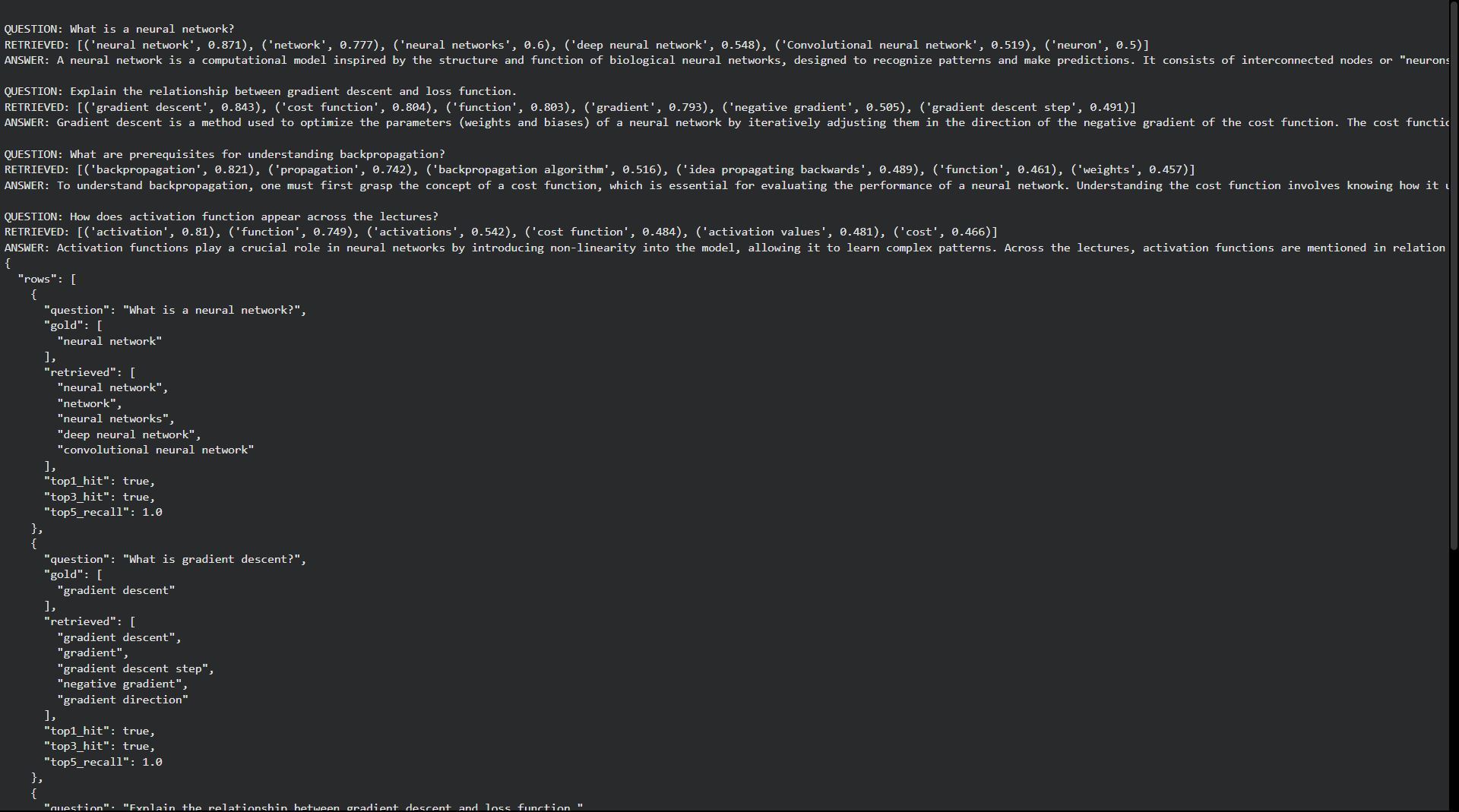}
  \caption{Representative questions, retrieved concepts with relevance scores, generated answers, and retrieval-evaluation records.}
  \label{fig:qaexamples}
\end{figure*}

Fig.~\ref{fig:qaexamples} shows representative retrieval and answer outputs for definition, relation, prerequisite, and cross-lecture questions. The examples illustrate how the system ranks relevant graph concepts before generating an answer, while the accompanying evaluation records whether the gold concepts occur among the highest-ranked results.

\begin{table}[t]
\caption{Starter Retrieval Evaluation}
\label{tab:retrieval}
\centering
\small
\begin{tabularx}{\columnwidth}{@{}Xccc@{}}
\toprule
Question (gold concept) & Top-1 & Top-3 & R@5 \\
\midrule
What is a neural network? (neural network) & Yes & Yes & 1.00 \\
What is gradient descent? (gradient descent) & Yes & Yes & 1.00 \\
How do gradient descent and loss function relate? (both) & Yes & Yes & 1.00 \\
\bottomrule
\end{tabularx}
\end{table}

\begin{table}[t]
\caption{Qualitative Grounded-QA Behavior}
\label{tab:qualitative}
\centering
\small
\begin{tabularx}{\columnwidth}{@{}l l X@{}}
\toprule
Type & Example & Observed behavior \\
\midrule
Definition & Neural network & Ranked the target first; described layers, neurons, weights, biases, activations, and cost. \\
Relation & Gradient descent/loss & Retrieved both first and explained negative-gradient optimization. \\
Prerequisite & Backpropagation & Identified cost, weights, biases, propagation, and gradient descent. \\
Cross-lecture & Activation function & Linked activation values to layer-to-layer flow with lecture timestamps. \\
\bottomrule
\end{tabularx}
\end{table}

\section{Discussion}
Evidence grounding makes the graph inspectable: each retained mention records lecture, anchor, frame, timestamp, modality, quotation, and confidence; each edge records endpoints, type, and supporting evidence. This reduces the risk of a structurally plausible but unsupported VLM-generated graph. The 94.63\% non-empty OCR rate also indicates that recall-oriented anchors frequently capture visible educational content that transcript chunks omit.

Canonicalization merged repeated mentions effectively but remains incomplete. Variants such as \emph{weight/weights}, \emph{bias/biases}, and \emph{cost/loss function} demonstrate the tension between over-merging distinct ideas and fragmenting aliases. Endpoint coverage is therefore a critical diagnostic: 90.38\% shows that most validated relations matched the final node inventory after deduplication.

Perfect retrieval on three seeds only confirms that obvious concepts are reachable. A credible benchmark must include substantially more definition, relation, prerequisite, example, first-mention, temporal-evolution, visually grounded, and cross-lecture questions.

\section{Limitations and Future Work}
The dataset contains only three lectures from one series, and extraction lacks a manually annotated concept-and-relation gold standard. The three-query retrieval set cannot support statistical claims. Canonicalization leaves aliases and singular/plural duplicates; isolated or noisy nodes may encode generic terms, numeric labels, or visual artifacts. OCR and VLM accuracy depend on frame resolution, handwriting, transitions, and diagram complexity. Finally, answer generation may add unsupported background knowledge even when it is correct.

Future work will (1) annotate concept mentions, canonical entities, relations, evidence validity, and QA; (2) add domain-aware lemmatization, alias dictionaries, and merge blocklists; (3) prune low-evidence isolated nodes and enrich reliable central relations; (4) compare transcript-only RAG, transcript-plus-OCR RAG, ungrounded extraction, grounded extraction without images, and the full pipeline; (5) evaluate answer faithfulness, citation validity, and unsupported-answer rate; and (6) scale across courses and domains with potentially different relation ontologies.

\section{Conclusion}
This paper presented an evidence-grounded multimodal pipeline for turning lecture videos into a provenance-rich KG for multi-lecture reasoning. Across three neural-network lectures it processed 3,118 frames, 756 transcript segments, and 559 anchors; retained 1,022 concept and 312 relation mentions; and constructed 172 canonical nodes and 282 edges with 90.38\% endpoint coverage. Preliminary retrieval succeeded on three seed questions, but larger annotations, controlled ablations, stronger canonicalization, and rigorous faithfulness evaluation are required. The principal result is an auditable method that connects graph structure to explicit lecture evidence.

\noindent\begin{minipage}{\columnwidth}
\section*{Reproducibility}
\footnotesize
The final run is named \path{evidence_grounded_run_001}. Artifact generation uses \path{docs/paper_artifacts/generate_artifacts.py}, and the final notebook is \path{notebooks/Evidence_Grounded_Multimodal_KG_Construction.ipynb}. Large dataset artifacts are documented by the project and hosted at \url{https://huggingface.co/datasets/sahilfarib/evidence-grounded-multimodal-kg-multi-lecture-reasoning}.
\end{minipage}

\end{document}